\pdfoutput=1
\documentclass[letterpaper]{article} 
\usepackage{aaai25}
\usepackage{times}  
\usepackage{helvet}  
\usepackage{courier}  
\usepackage[hyphens]{url}  
\usepackage{graphicx} 
\usepackage{tabularx}
\usepackage{natbib}  
\usepackage{caption} 
\usepackage{algorithm}
\usepackage{algorithmic}
\usepackage{booktabs}
\usepackage{array}
\usepackage{subfig}
\usepackage{amsmath}

\usepackage[T1]{fontenc}
\usepackage[utf8]{inputenc}
\usepackage{newfloat}
\usepackage{listings}
\DeclareCaptionStyle{ruled}{labelfont=normalfont,labelsep=colon,strut=off} 
\floatstyle{ruled}
\newfloat{listing}{tb}{lst}{}
\floatname{listing}{Listing}
\title{Order Matters: Exploring Order Sensitivity in Multimodal Large Language Models}
\author {
    Zhijie Tan\textsuperscript{\rm 1},
    Xu Chu\textsuperscript{\rm 1},
    Weiping Li\textsuperscript{\rm 1},
    Tong Mo\textsuperscript{\rm 1}
}
\affiliations {
    \textsuperscript{\rm 1}School of Software and Microelectronics, Peking University\\
    besttangent@stu.pku.edu.cn, chuxu@stu.pku.edu.cn, wpli@ss.pku.edu.cn, motong@ss.pku.edu.cn
}

\usepackage{bibentry}

\begin{document}

\maketitle

\begin{abstract}
Multimodal Large Language Models (MLLMs) utilize multimodal contexts consisting of text, images, or videos to solve various multimodal tasks. However, we find that changing the order of multimodal input can cause the model's performance to fluctuate between advanced performance and random guessing. This phenomenon exists in both single-modality (text-only or image-only) and mixed-modality (image-text-pair) contexts. Furthermore, we demonstrate that popular MLLMs pay special attention to certain multimodal context positions, particularly the beginning and end. Leveraging this special attention, we place key video frames and important image/text content in special positions within the context and submit them to the MLLM for inference. This method results in average performance gains of 14.7\% for video-caption matching and 17.8\% for visual question answering tasks. Additionally, we propose a new metric, Position-Invariant Accuracy (PIA), to address order bias in MLLM evaluation. Our research findings contribute to a better understanding of Multi-Modal In-Context Learning (MMICL) and provide practical strategies for enhancing MLLM performance without increasing computational costs.
\end{abstract}

%

\section{Introduction}

The recent success of Large Language Models (LLMs)~\cite{Zhao2023ASO,Brown2020LanguageMA,Touvron2023LLaMAOA,qwen,Xi2024TrainingLL} has driven researchers to explore their capabilities in tackling multimodal tasks. By aligning image features with text embeddings, researchers integrate visual inputs into LLMs, developing Multimodal Large Language Models (MLLMs)~\cite{Yin2023ASO,Li2023BLIP2BL,Alayrac2022FlamingoAV,Li2023VideoChatCV,Beyer2024PaliGemmaAV}. These models inherit the exceptional abilities of LLMs in In-Context Learning (ICL) and have demonstrated significant performance in understanding and reasoning with images and videos.

A core issue in In-Context Learning for LLMs is prompt order. It has been shown in LLMs that there are better prompt orders (which vary between different models or tasks). When ordered correctly, the models perform well, whereas other orders result in performance close to random~\cite{Lu2021FantasticallyOP}. The observation of prompt order sensitivity in LLMs raises two compelling questions. Question \uppercase\expandafter{\romannumeral1}: ``Does order sensitivity also exist in MLLMs?" Furthermore, another question that needs to be discussed is, Question \uppercase\expandafter{\romannumeral2}: ``What kind of order is good for MLLM's performance?" Current research and benchmark studies~\cite{shi2024judging,chen2024mllm,Liu2023MMBenchIY} focus on addressing a sub-question of Question \uppercase\expandafter{\romannumeral1}: ``Does \textit{text-only} order sensitivity also exist in MLLMs?". These works overlook the potential influence of image order and situations
where both image and text orders are altered. 


\begin{figure}[t]
\centering
\includegraphics[width=0.49\textwidth]{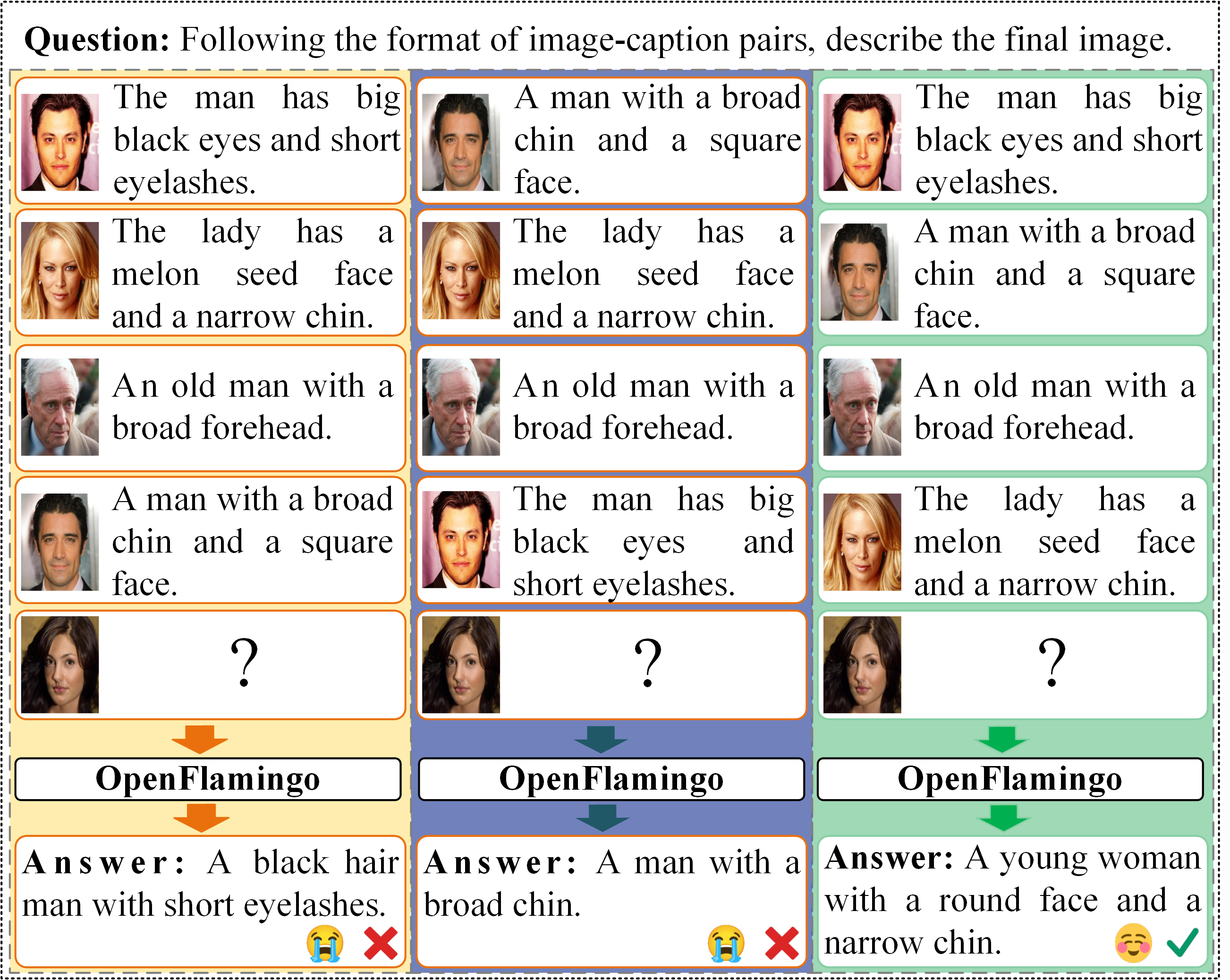} 
\caption{Certain multimodal context orders elicit accurate responses, while others lead to erroneous outputs. The green demonstration indicates good order, while the yellow and blue demonstrations indicate bad order.}
\label{introexp}
\end{figure}

\begin{figure*}[ht]
    \centering
    \subfloat[Text-only: Provide 4 images (different distinct views of the same facial image) and 4 corresponding captions, with the order of the captions randomly shuffled. The task for the model is to generate new captions.]
    {
        \includegraphics[width=0.3\textwidth]{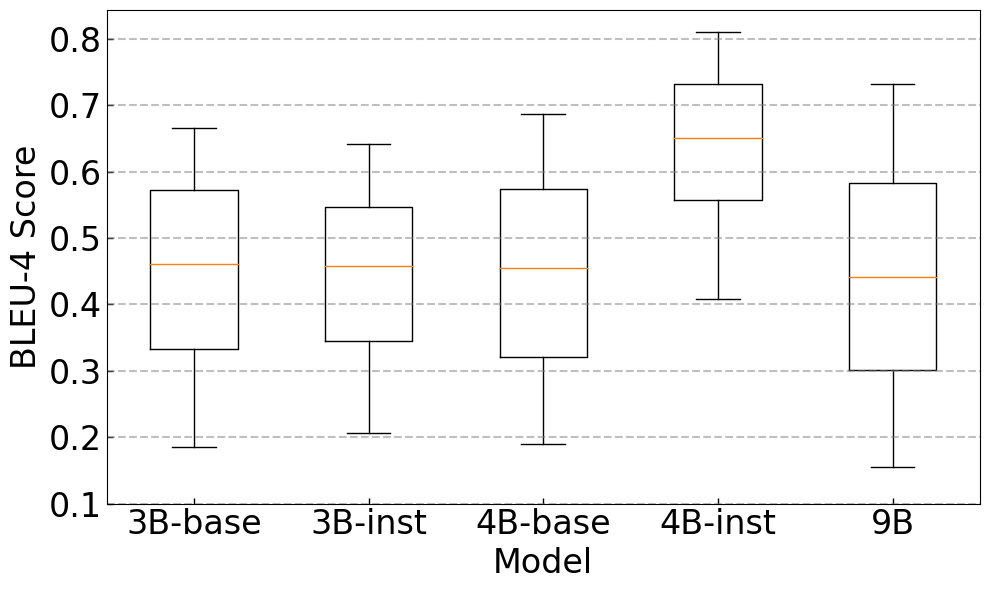}
        \label{fig:subfig1}
    }
    \hfill
    \subfloat[Image-only: Provide 4 images (different distinct views of the same facial image) and 4 corresponding captions, with the order of the images randomly shuffled. The task for the model is to generate new captions.]
    {
        \includegraphics[width=0.3\textwidth]{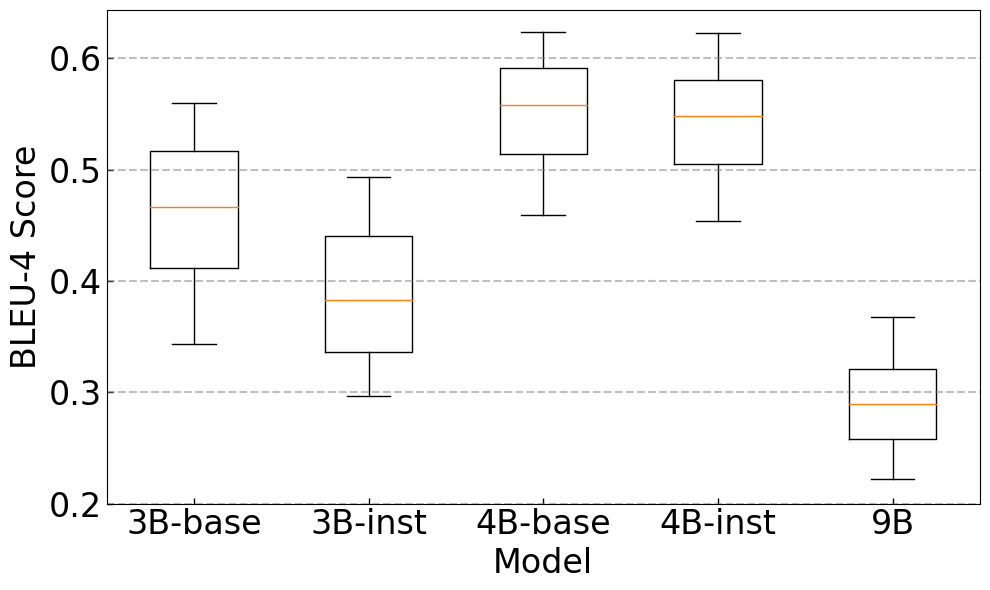}
        \label{fig:subfig2}
    }
    \hfill
    \subfloat[Image-text-pair: Provide 4 images (facial images of 4 different persons) and 4 corresponding captions, with the order of the image-caption pairs randomly shuffled. The task for the model is to generate captions for a new facial image.]
    {
        \includegraphics[width=0.3\textwidth]{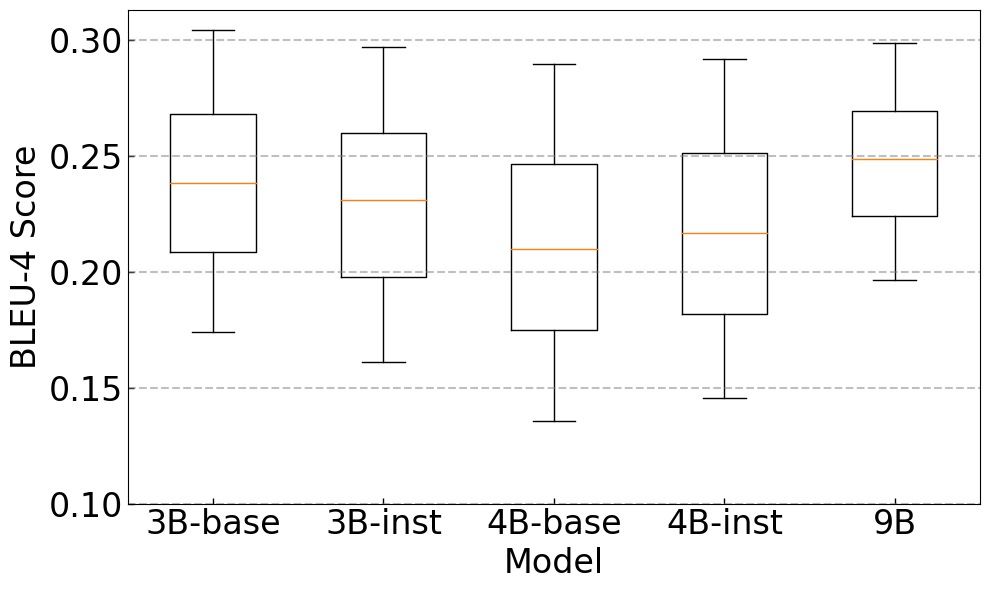}
        \label{fig:subfig3}
    }
    
    \caption{The performance of OpenFlamingo \cite{Awadalla2023OpenFlamingoAO} (3B-base, 3B-instruct, 4B-base, 4B-instruct, 9B) with different multimodal context orders in the image captioning task.}
    \label{fig:openflamingo-boxplot}
\end{figure*}

\begin{figure}[tb]
\centering
\includegraphics[width=0.45\textwidth]{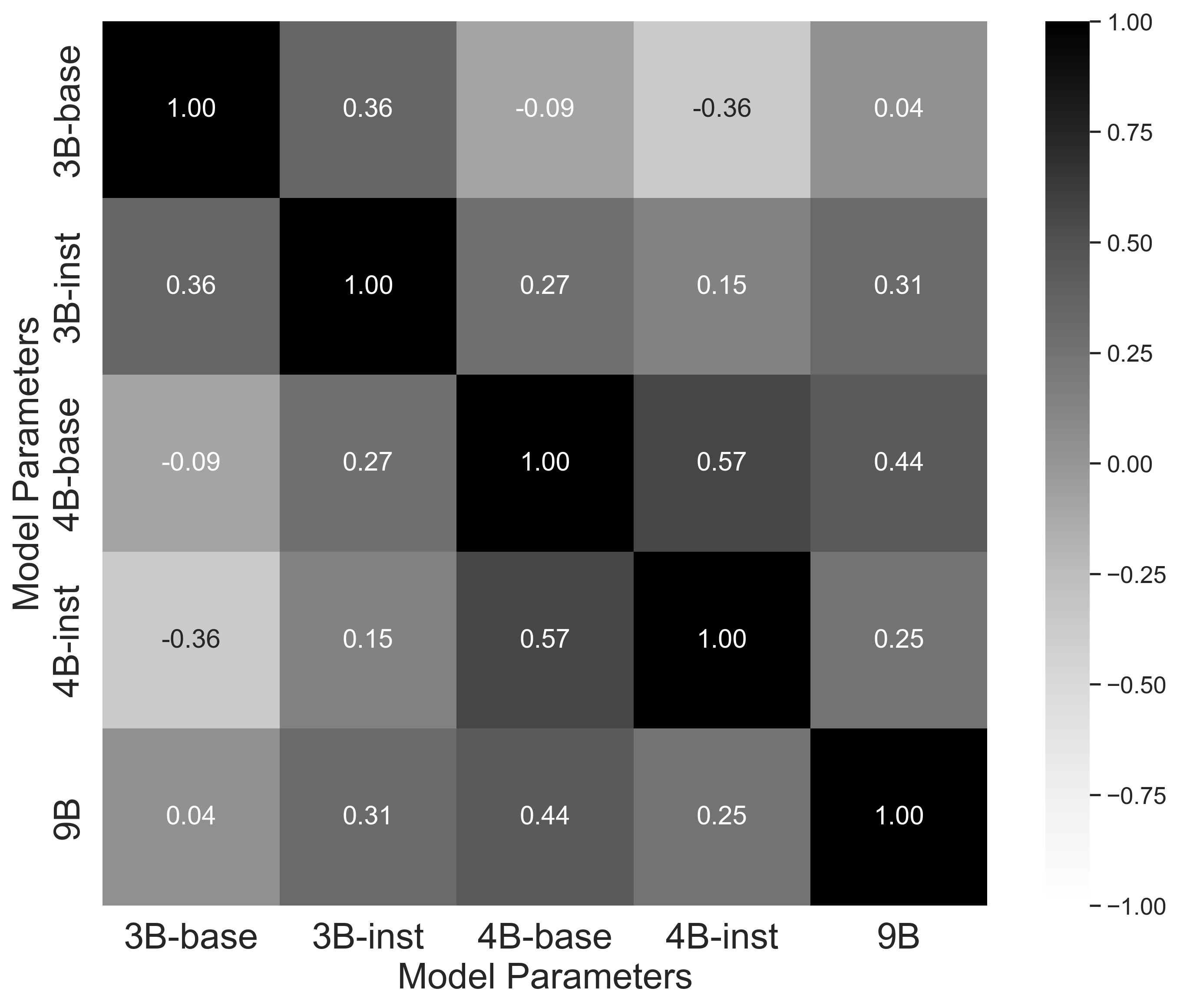} 
\caption{Spearman correlation coefficients of multimodal context order performance across OpenFlamingo models with varying parameter scales.}
\label{spearman}
\end{figure}

\begin{figure}[t]
\centering
\includegraphics[width=0.45\textwidth]{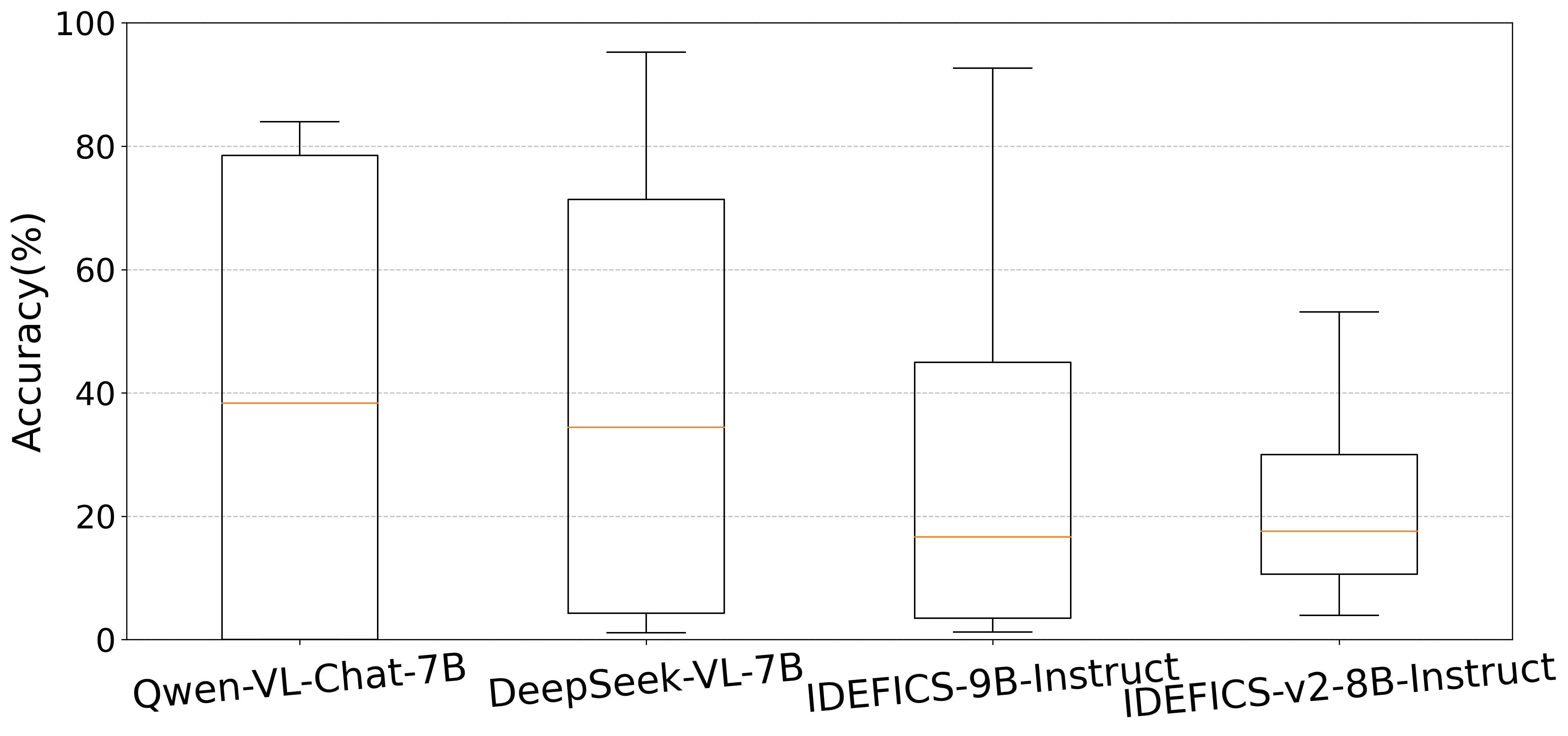} 
\caption{Performance of Qwen-VL-Chat-7B~\cite{Qwen-VL}, DeepSeek-VL-7B~\cite{lu2024deepseekvl}, IDEFICS-9B-Instruct~\cite{laurencon2023obelics} and IDEFICS-v2-8B-Instruct~\cite{Laurenon2024WhatMW} on image captioning task for modified COCO dataset.}
\label{4models}
\end{figure}

To fully answer Question \uppercase\expandafter{\romannumeral1}, we design the following experiments. As shown in Figure \ref{introexp}, CelebAText-HQ~\cite{Sun2021MulticaptionTS} is a facial image captioning dataset. We design a 4-shot image-text prompt and ask OpenFlamingo\cite{Awadalla2023OpenFlamingoAO}, a widely studied MLLM for multi-image input tasks~\cite{Yang2023ExploringDI, chen2024visual, schlarmann2023adversarial}, to generate a caption for a new facial image. When the image-text-pairs are arranged in good order (as illustrated in green in Figure \ref{introexp}), the model can generate essentially accurate responses. However, bad orders cause the model's responses to approach randomness (as shown in yellow and blue in Figure \ref{introexp}). Furthermore, as depicted in Figure \ref{fig:openflamingo-boxplot}, we demonstrate that altering the prompt order of text-only, image-only, or image-text-pair all impact the model's performance in image captioning tasks. For text-only and image-only tasks, we employ HFGI3D~\cite{Xie2022Highfidelity3G} to generate four distinct views of each facial image as visual prompts. The detail of text-only and image-only tasks can be seen in Figure~\ref{intro_text_only} and Figure~\ref{intro_image_only} of Appendix A. The results indicate that both single-modality context order and mixed-modality context order significantly influence the model's context learning performance. Moreover, we discover that high-performing prompts are not transferable across models. As shown in Figure \ref{spearman}, we utilize all 24 possible orderings of the 4-shot image-text prompts in the image-text-pair task. We then generate predictions using each prompt across different parameter versions of OpenFlamingo~\cite{Awadalla2023OpenFlamingoAO} and calculate pairwise Spearman rank correlation coefficients between the scores. The low correlations observed, such as \textit{-0.09} between OpenFlamingo-3B-base and OpenFlamingo-4B-base, suggest that orderings beneficial for one model do not necessarily yield good performance in another. The degree of order sensitivity also varies among models. Figure \ref{4models} presents results from an image-caption matching task using modified COCO~\cite{Lin2014MicrosoftCC} dataset (for a detailed introduction, refer to the image-text-pair task in Figure~\ref{table1} of Appendix B). It reveals that Qwen-VL-Chat-7B~\cite{Qwen-VL} and DeepSeek-VL-7B~\cite{lu2024deepseekvl} exhibit stronger order sensitivity for image-text-pairs compared to IDEFICS-9B-Instruct ~\cite{laurencon2023obelics} and IDEFICS-v2-8B-Instruct~\cite{Laurenon2024WhatMW}. In summary, regarding Questions \uppercase\expandafter{\romannumeral1}, we believe that order sensitivity also exists in MLLMs, and changing the order of text-only, image-only, and image-text-pairs in the prompt all affect model performance. The magnitude of this impact varies for different models and different tasks.

\begin{figure}[t]
\centering
\includegraphics[width=0.45\textwidth]{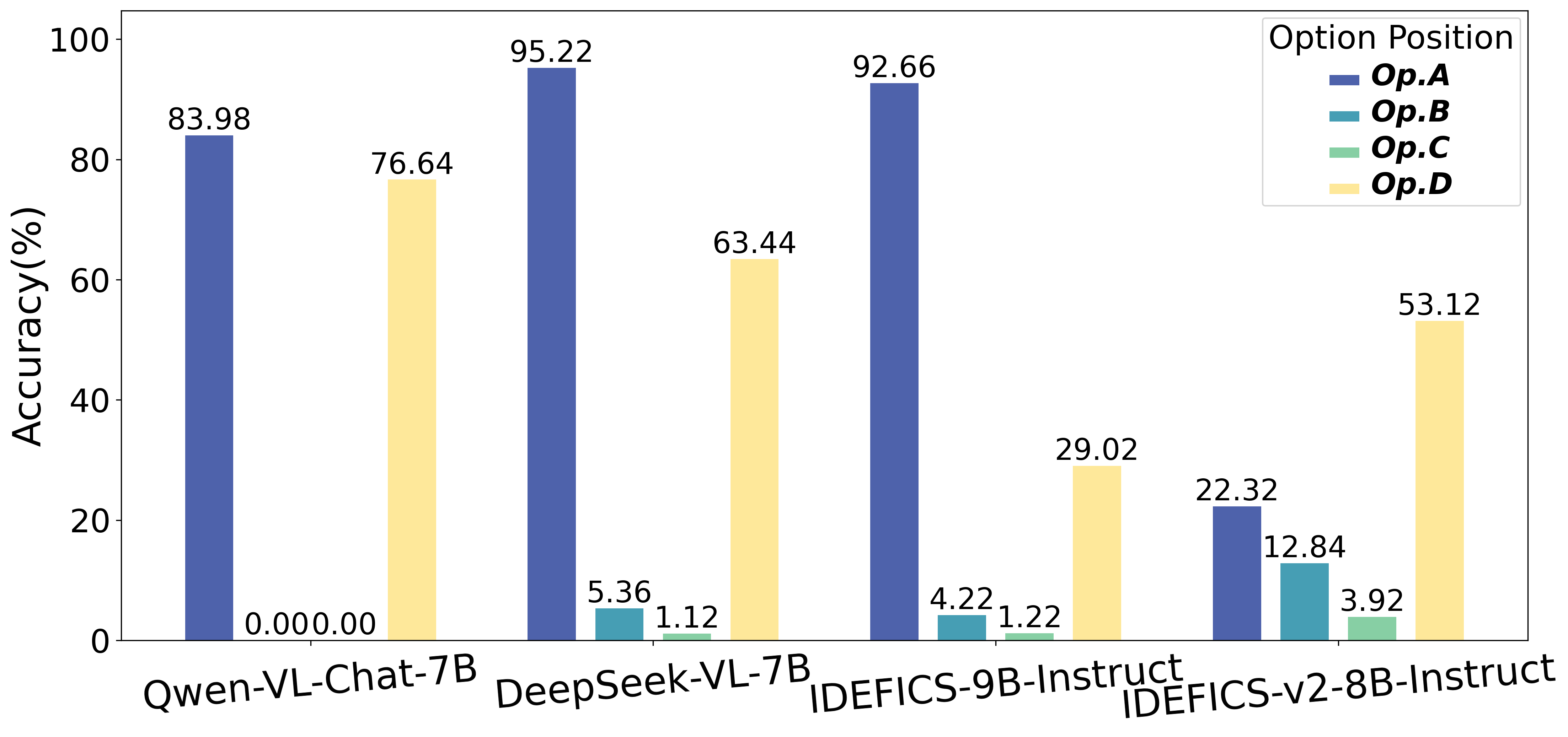} 
\caption{Performance of Qwen-VL-Chat-7B, DeepSeek-VL-7B, IDEFICS-9B-Instruct and IDEFICS-v2-8B-Instruct on image captioning task for modified COCO dataset, with different option positions.}
\label{4models_grouped}
\end{figure}

Moreover, for Question \uppercase\expandafter{\romannumeral2}: ``What kind of order is good for MLLM's performance?" Previous research on prompt order in LLMs~\cite{Lu2021FantasticallyOP,Wu2022SelfAdaptiveIL,Wang2023LargeLM,Li2023SplitAM,Xu2023RetrievalML,Song2024HierarchicalCM} and our experiments on MLLMs both indicate that no universally optimal prompt order exists that generalizes well across different models and tasks. While many studies explore optimal prompt order~\cite{Lu2021FantasticallyOP, Wu2022SelfAdaptiveIL,Yoo2021GPT3MixLL}, they are often specific to certain modalities or tasks, making it difficult to extend these approaches to multimodal or universal tasks. Interestingly, considering that an order is composed of arrangements of different positions, we discover that popular MLLMs pay special attention to certain contextual positions, particularly the \textbf{beginning} and \textbf{end} of the input. As shown in Figure \ref{4models_grouped}, in the image-caption matching task using the modified COCO dataset, placing the correct answer (image-text-pair) at the first and last options (\textbf{\textit{Op.A}} and \textbf{\textit{Op.D}}) results in the highest accuracy for all four models. Conversely, placing the correct answer in the middle options (\textbf{\textit{Op.B}} and \textbf{\textit{Op.C}}) causes a significant drop in accuracy, with Qwen-VL-Chat-7B~\cite{Qwen-VL} unable to provide the correct answer at all. This answers Question \uppercase\expandafter{\romannumeral2}: The order formed by placing important content at the beginning or end of the multimodal context is beneficial to the performance of popular MLLMs.

To utilize the order sensitivity to improve performance, we design two order-sensitive tasks for Multi-Modal In-Context Learning (MMICL): a video-caption matching task and a visual question answering task with Retrieval-Augmented Generation (RAG)~\cite{lewis2020retrieval, Chen2022MuRAGMR, caffagni2024wiki}. In these experiments, we place keyframes of the video or important prompts in special positions within the multimodal context, demonstrating that this approach improves the accuracy of the model. This indicates that special positions in MLLMs contain rewards that can enhance the model's capabilities. Additionally, we propose a new MLLM metric: Position-Invariant Accuracy (PIA). This metric aims to eliminate unfairness caused by order bias in models. PIA operates by cyclically placing the correct option in each position, then counting the number of times the model selects each option and the number of correct selections among them. Through weighted scoring, the essence of PIA is to reduce the bias introduced by the model's tendency to favor certain options. We apply this metric to evaluate several widely studied and used MLLMs using the 4-option MMBench dataset~\cite{Liu2023MMBenchIY}.

Our contributions are as follows:
\begin{itemize}
    \item We demonstrate that order sensitivity exists in MLLMs, and changing the order of prompts can significantly affect the performance of MLLMs.
    \item We find that orders composed of some special positions are beneficial for MLLMs. Popular MLLMs pay special attention to the beginning and end of multimodal contexts. Placing important content in these positions can improve performance without additional computation.
    \item We prove that our finding on order sensitivity can be beneficial for order-dependent tasks such as video understanding and visual question answering, as demonstrated by our experiments on video-caption matching and visual question answering with RAG. The results show that using rewards for special positions in the context of understanding keyframes in videos and important prompts in RAG can improve model performance.
    \item We propose Position-Invariant Accuracy (PIA), a new MLLM metric designed to eliminate unfairness caused by order biases in MLLMs.
\end{itemize}

\section{Related Work}
\subsection{Multimodal Large Language Model}
Multimodal Large Language Model (MLLM) consist of pre-trained modality encoders, Large Language Models (LLMs), and modality alignment interfaces. Encoders like CLIP~\cite{Radford2021LearningTV} use contrastive learning to align images and text for image-based tasks. Pre-trained LLMs~\cite{Ouyang2022TrainingLM,Bai2022ConstitutionalAH,Touvron2023LLaMAOA}, through extensive training on corpora, embed rich knowledge and exhibit strong generalization and reasoning capabilities. Modality alignment interfaces combine image-focused modality encoders with LLMs, which only understand text, to perform various multimodal tasks. The mainstream approach introduces learnable connectors between pre-trained modality encoders and LLMs. These connectors include learnable query tokens (e.g., BLIP-2~\cite{Li2023BLIP2BL}), MLP (e.g., LLAVA~\cite{liu2023llava}), linear layer (e.g., PaliGemma~\cite{Beyer2024PaliGemmaAV}), and the insertion of additional learnable modules (e.g., Flamingo~\cite{Alayrac2022FlamingoAV}, LLAMa-Adapter~\cite{Zhang2023LLaMAAdapterEF}, and LLaVA-TokenPacker~\cite{Li2024TokenPackerEV}). 
\subsection{Promt Learning in MLLM}
Prompt learning can be divided into hard prompts (discrete prompts)~\cite{Wen2023HardPM,choi2024hard} and soft prompts (continuous prompts)~\cite{qin2021learning,Wu2023InfoPromptIS}. Hard prompts are typically composed of natural language words, making them easily understandable by humans. In contrast, soft prompts are usually learnable continuous vectors. Soft prompts are widely used in multimodal prompt learning, existing as continuous vectors in the embedding space of LLMs and being automatically learned during the fine-tuning process. CoOp~\cite{Zhou2021LearningTP} fine-tunes CLIP by optimizing a set of continuous prompt vectors on its language branch to achieve a few-shot transfer. Additionally, many studies focus on optimizing prompts for single modalities in MLLM~\cite{Bahng2022VisualPM,Ju2021PromptingVM,Lu2022PromptDL}. Unlike these works, Maple~\cite{Khattak2022MaPLeMP} proposes a multimodal joint prompting method that couples prompts from both the language and image branches to fine-tune CLIP, achieving better results than single-modality prompting. ViLT-CLIP~\cite{wang2024vilt} introduces a method to reduce the discrepancy between hand-crafted prompts and learnable prompts, mitigating the forgetting of essential video scenarios and achieving the transfer of image-based CLIP to the video domain. ControlMLLM~\cite{wu2024controlmllm} dynamically adjusts visual-textual token interactions during inference, enabling precise regional descriptions and reasoning.
\subsection{Sensitivity of LLMs to Input Order}
Jiang et al.\cite{Jiang2019HowCW} find that LLMs are highly sensitive to the order of prompts in zero-shot and few-shot settings. Kumar and Talukdar explore finding the optimal order of training examples as enhanced prompts and learning delimiter tokens between prompts to further improve performance \cite{Kumar2021ReorderingEH}. Lu et al.\cite{Lu2021FantasticallyOP} discover that the order in which response prompts are given to LLMs plays a crucial role in their performance and propose an entropy-based method to score different candidate orderings. Similarly, inspired by the learning compressed perspectives, Wu et al.\cite{Wu2022SelfAdaptiveIL} optimize the order by minimizing the encoding length required to compress and transmit test labels. Ma et al.\cite{Ma2023FairnessguidedFP} address the challenge of order in few-shot learning from the perspective of prediction bias. They introduce a greedy search strategy to identify the optimal prompt order. Zhang et al.~\cite{Zhang2024BatchICLEE} tackle the order dependency in context learning by converting the n-shot problem into n parallel 1-shot tasks.

\begin{figure}[t]
\centering
\includegraphics[width=0.49\textwidth]{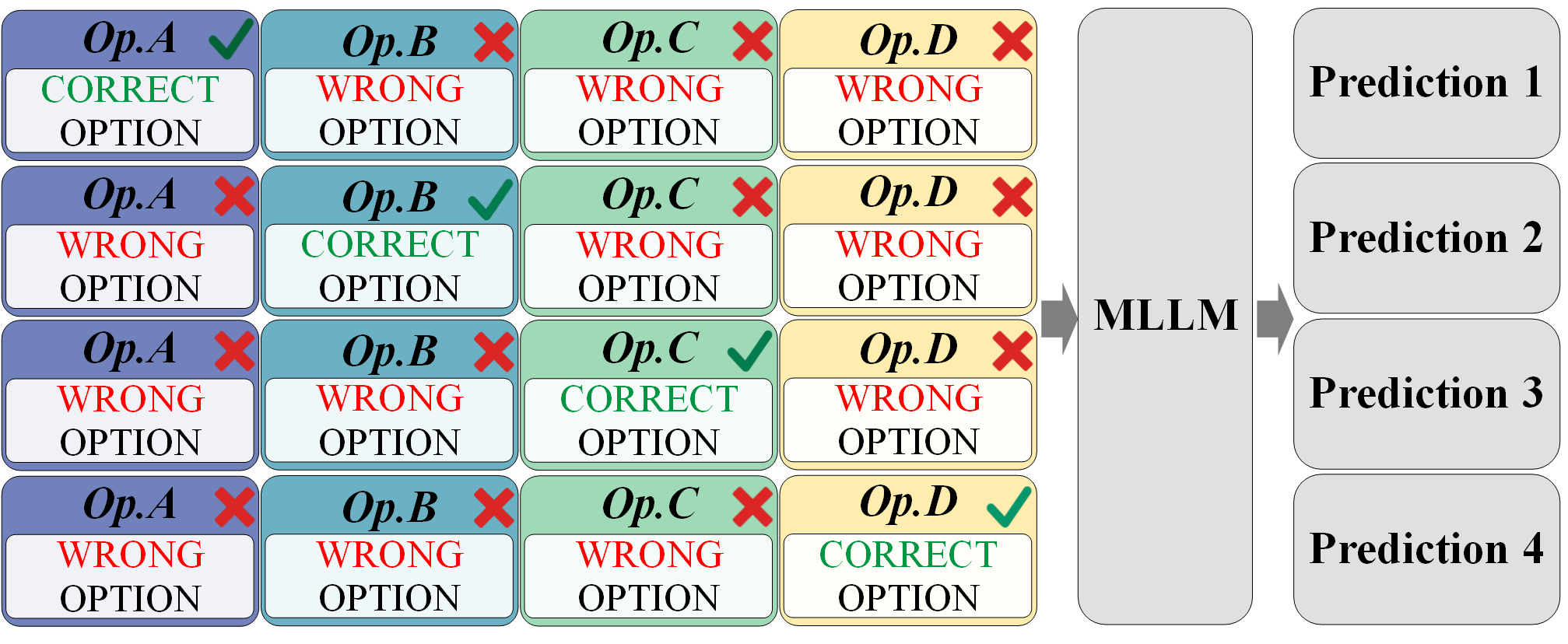} 
\caption{Evaluating the model's performance by varying the position of the correct option within the multimodal context.}
\label{order_sample}
\end{figure}

\section{Order Sensitivity}
Previous research~\cite{Lu2021FantasticallyOP, Wu2022SelfAdaptiveIL,Yoo2021GPT3MixLL} focuses on finding optimal orders for specific modalities and tasks. However, for a broader range of modalities and tasks, a general brute-force enumeration method for searching the optimal order requires unacceptable time consumption. Considering that an order is determined by the arrangement of different positions, we observe in extensive experiments that popular MLLMs focus on specific positions within multimodal contexts. Therefore, our research focuses on utilizing these specific positions to construct better orders. In this section, we design three tasks to explore MLLM's preferences for text-only positions, image-only positions, and image-text-pair positions. As shown in Figure~\ref{order_sample}, by altering the position of the correct option in the multimodal context and counting the times the model correctly answers questions at different positions, we assess the model's position preference. Examples and descriptions of the three tasks are shown in Figure~\ref{table1} of Appendix B.


\begin{figure*}[t]
\centering
\includegraphics[width=0.95\textwidth]{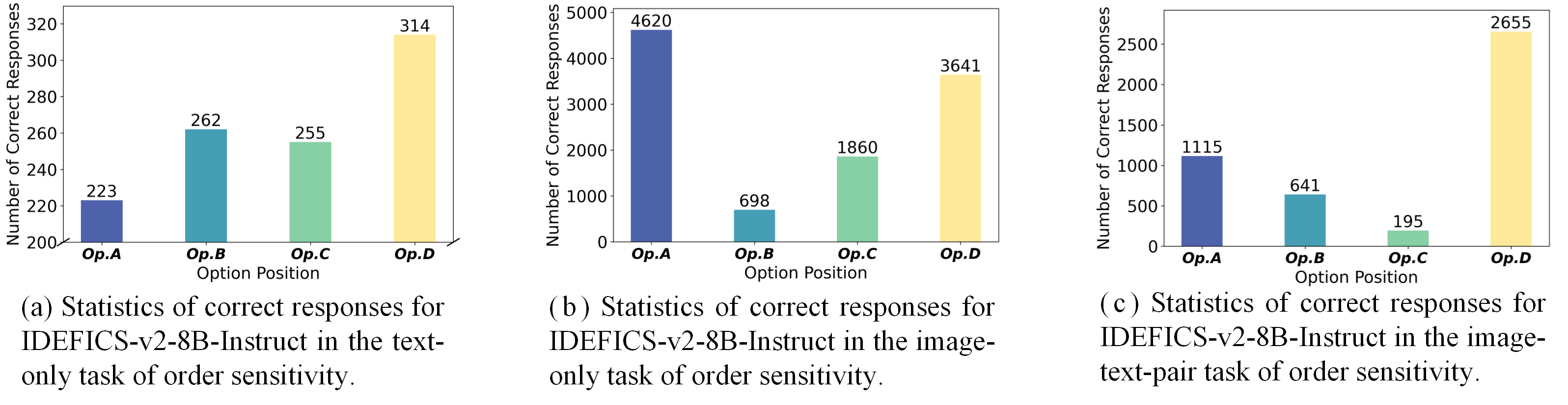} 
\caption{Statistics of correct responses for IDEFICS-v2-8B-Instruct~\cite{Laurenon2024WhatMW} in order sensitivity tasks.}
\label{tasks}
\end{figure*}

    

\subsection{Text-only Order Sensitivity}
We use the multiple-choice question-answering dataset MMBench~\cite{Liu2023MMBenchIY} to evaluate MLLM's order preferences in text-only. Each question in this dataset contains an image and a question about the image, along with up to four text candidate options, with only one correct answer. For our experiments, we select questions with four candidate options and rotate the correct answer through the \textbf{\textit{Op.A}}, \textbf{\textit{Op.B}}, \textbf{\textit{Op.C}}, and \textbf{\textit{Op.D}} positions.  The implementation details are as follows. Given question $Q$ and the context input:
\begin{equation}
    S_\text{text-only} = \{I; T_1; T_2; \ldots; T_n\},
\end{equation}
where $I$ represents the image on which question $Q$ depends, and $T_1; T_2; \ldots; T_n$ represent the $n$ text options, the model is required to answer question $Q$ based on context $S_\text{text-only}$ and choose the correct option from $T_1; T_2; \ldots; T_n$.

In our experimental setup, we set $n=4$ and uniquely label the position of each text option as $O_1 = (\textbf{\textit{Op.A}}, T_1)$; $O_2 = (\textbf{\textit{Op.B}}, T_2)$; $O_3 = (\textbf{\textit{Op.C}}, T_3)$; $O_4 = (\textbf{\textit{Op.D}}, T_4)$. We designate $\textbf{\textit{Op.A}}$ as the option representing the first position and $\textbf{\textit{Op.D}}$ as the option representing the last position. The context input is updated to:
\begin{equation}
    S_\text{text-only} = \{I; O_1; O_2; O_3; O_4\}.
\end{equation}

To explore the text-only order sensitivity of MLLM in In-Context Learning, we exchange the correct answer $T_\text{true}$ with $T_1$, $T_2$, $T_3$, and $T_4$ respectively, sequentially generating four variants of each question, and input them into the model in turn to obtain answers.

We conduct experiments on over 4,000 questions using IDEFICS-v2-8B-Instruct~\cite{Laurenon2024WhatMW}. To more clearly illustrate the impact of order on model performance, for each question, we exclude from the statistical data the results of questions that the model answers correctly in all four variants. We then tally the positions of the correct option when the model responds accurately. The results, shown in Figure~\ref{tasks}(a), indicate that the model achieves the highest accuracy when the correct option is in position \textbf{\textit{Op.D}} (i.e., the last option).

\subsection{Image-only Order Sensitivity}
We use the COCO dataset~\cite{Lin2014MicrosoftCC} to create a multiple-choice image-caption matching task to evaluate MLLM's order preferences in image-only. We select 5000 images with corresponding captions as correct answers. For each question, we add three random images as alternatives. The task is to choose the image best matches the caption, with one correct answer. The correct answer in each question is sequentially swapped among the \textbf{\textit{Op.A}}, \textbf{\textit{Op.B}}, \textbf{\textit{Op.C}}, and \textbf{\textit{Op.D}} options. The implementation is as follows. Given question $Q$ and the context input:
\begin{equation}
    S_\text{image-only} = \{T; I_1; I_2; \ldots; I_n\},
\end{equation}
where $T$ represents the caption text, and $I_1; I_2; \ldots; I_n$ represent the $n$ image options. The model is required to choose the image option that best matches the caption $T$ from $I_1; I_2; \ldots; I_n$.

In our experimental setup, we set $n=4$ and uniquely label the position of each image option as $O_1 = (\textbf{\textit{Op.A}}, I_1)$; $O_2 = (\textbf{\textit{Op.B}}, I_2)$; $O_3 = (\textbf{\textit{Op.C}}, I_3)$; $O_4 = (\textbf{\textit{Op.D}}, I_4)$. We designate $\textbf{\textit{Op.A}}$ as the option representing the first position and $\textbf{\textit{Op.D}}$ as the option representing the last position. The context input is updated to:
\begin{equation}
    S_\text{image-only} = \{T; O_1; O_2; O_3; O_4\}.
\end{equation}

To explore the image-only order sensitivity of MLLM in In-Context Learning, we exchange the correct answer $I_\text{true}$ with each of $I_1; I_2; I_3; I_4$ respectively, sequentially generating four variants of each question, and input them into the model in turn to obtain answers.

We conduct experiments on 5,000 questions using IDEFICS-v2-8B-Instruct~\cite{Laurenon2024WhatMW}. To more clearly illustrate the impact of order on model performance, for each question, we exclude from the statistical data the results of questions that the model answers correctly in all four variants. We then tally the positions of the correct option when the model responds accurately. The results, shown in Figure~\ref{tasks}(b), indicate that the model's accuracy is highest when the correct answer is in position \textbf{\textit{Op.A}} or \textbf{\textit{Op.D}} (the first or last option).

\subsection{Mixed-modality Order Sensitivity}\label{image-text-pair}
We utilize the COCO dataset~\cite{Lin2014MicrosoftCC} to construct a multiple-choice image-caption matching task to evaluate MLLM's order preferences for mixed modalities (image-text-pairs). We select 5000 image-caption pairs as correct answers. For each question, we add three random image-caption pairs as alternatives. The task is to choose the best matching pair, with one correct answer. The correct option in each question is sequentially swapped among the \textbf{\textit{Op.A}}, \textbf{\textit{Op.B}}, \textbf{\textit{Op.C}}, and \textbf{\textit{Op.D}} options. The implementation is as follows. Given question $Q$ and the context input:
\begin{equation}
    S_\text{image-text-pair} = \{(I_1, T_1); (I_2, T_2); \ldots; (I_n, T_n)\},
\end{equation}
where $(I_1, T_1); (I_2, T_2); \ldots; (I_n, T_n)$ represent $n$ image-text-pairs. The model is required to choose the most matching image and caption pair from $(I_1, T_1); (I_2, T_2); \ldots; (I_n, T_n)$.

In our experimental setup, we set $n=4$ and label each image-text-pair option with a unique position, namely $O_1=(\textbf{\textit{Op.A}}, (I_1, T_1))$; $O_2=(\textbf{\textit{Op.B}}, (I_2, T_2))$; $O_3=(\textbf{\textit{Op.C}}, (I_3, T_3))$; $O_4=(\textbf{\textit{Op.D}}, (I_4, T_4))$. We designate $\textbf{\textit{Op.A}}$ as the option representing the first position and $\textbf{\textit{Op.D}}$ as the option representing the last position. The context input is updated to:
\begin{equation}
    S_\text{image-text-pair} = \{O_1; O_2; O_3; O_4\}.
\end{equation}

To explore the mixed-modality order sensitivity of MLLM in In-Context Learning, we exchange the correct answer $(I, T)_\text{true}$ with each of $(I_1, T_1); (I_2, T_2); (I_3, T_3); (I_4, T_4)$ respectively, sequentially generating four variants of each question, and input them into the model in turn to obtain answers.

We experiment with 5,000 questions using IDEFICS-v2-8B-Instruct~\cite{Laurenon2024WhatMW}. To more clearly illustrate the impact of order on model performance, for each question, we exclude from the statistical data the results of questions that the model answers correctly in all four variants. We then tally the positions of the correct option when the model responds accurately. Figure~\ref{tasks}(c) shows the model's accuracy is highest when the correct answer is positioned in \textbf{\textit{Op.A}} or \textbf{\textit{Op.D}} (first or last option).

\begin{table}[htb]
    \centering
    \begin{tabular}{c|c|c}
    \toprule[1.5pt]
        Model & Parameters & Tokens Per Image \\ \midrule
        DeepSeek-VL & 7B & 576 \\ 
        LLaVa-NeXT & 7B/13B/34B & 2880 \\ 
        IDEFICS-v2 & 8B & 320 \\ 
        MM1-Chat & 7B/30B & 720 \\ \bottomrule[1.5pt]
    \end{tabular}
    \caption{Tokens per image in DeepSeek-VL~\cite{lu2024deepseekvl}, LLaVa-NeXT~\cite{liu2023llava}, IDEFICS-v2~\cite{Laurenon2024WhatMW}, and MM1-Chat~\cite{McKinzie2024MM1MA}.}
    \label{tokens}
\end{table}
\subsection{Analysis}
The experimental results indicate that MLLMs prefer the beginning and end of the multimodal context. When the correct answer is positioned at the beginning or end, the model's chances of accurately answering the question increase. Even if the model cannot answer the question correctly, it tends to rely on the beginning and end of the context for its response. This phenomenon is more pronounced in image-only or mixed-modality tasks than in text-only tasks. The token length of encoded images is closer to that of long texts, as shown in Table~\ref{tokens}. This suggests that order bias may relate to the context length, with longer text lengths potentially amplifying this effect.

These results inspire us to consider: Would tasks influenced by order, such as video understanding and visual question answering (when visual prompts have an order), benefit from orders composed of these special positions? In the next section, we design two experiments: a video-caption matching task and a visual question answering task with RAG, to demonstrate the impact of prompt orders composed of special positions on these tasks.

\section{Experiment}
\subsection{Settings}
We use Video-LLaVA~\cite{Lin2023VideoLLaVALU} for the video-caption matching task and IDEFICS-v2-8B-Instruct~\cite{Laurenon2024WhatMW} for the visual question answering task with Retrieval-Augmented Generation (RAG). The experimental setup can be seen in Appendix C.


\subsection{Dataset}
\textbf{Video-Caption Matching.} Task construction is shown in Figure~\ref{exp_video} of Appendix C. The  MVBench~\cite{Li2023MVBenchAC} dataset serves as the foundation. MVBench is a commonly used dataset for MLLM instruction fine-tuning~\cite{Li2023VideoChatCV,Zhang2023VideoLLaMAAI}, containing multiple tasks and video collections. We focus on the fine-grained pose video set, comprising 200 fine-grained action video clips with corresponding textual captions. Our selection criteria include action lengths between 2-3 seconds (135 video clips), which we extend to create new 10-second videos. In each new video, we designate the original 2-3 second clip as the key action and randomly sample other action clips, concatenating them sequentially until reaching 10 seconds. Each video pairs with a caption-matching question: ``Is anyone in the video performing the following action: {action}?" where ``{action}" represents the textual caption of the key action. The model needs to answer with ``Yes" or ``No".

The video frames corresponding to the key action are referred to as \textit{Key Frames} (as the frames within the green boxes in Figure~\ref{exp_video}). For each video, we construct three comparative groups: we move \textit{Key Frames} to the front, middle, and back, respectively, resulting in three new videos with different arrangements of \textit{Key Frames}. The corresponding question and answer for each video remain unchanged.

\textbf{Visual Question Answering with RAG.} Task construction can be seen in Figure~\ref{exp_rag} of Appendix C. The multiple-choice question-answering dataset MMbench~\cite{Liu2023MMBenchIY} serves as the foundation for our construction. Each question in MMbench includes an image and a question about the image, along with up to 4 textual candidate options, with only one correct answer. To simulate RAG results, for each question, we refer to the image corresponding to the question as the \textit{Question-relevant image} (as the image within the green boxes in Figure~\ref{exp_rag}), and randomly select 3 other questions, obtaining their corresponding images. We combine these 4 images to form a RAG image set, in which only 1 image is relevant to the question, namely the \textit{Question-relevant image}. In our newly constructed dataset, each entry includes a RAG image set, a question about the \textit{Question-relevant image}, and four text candidate options.

For each question, we create four comparative groups by positioning the \textit{Question-relevant image} in the first, second, third, and fourth positions within the RAG image set, respectively. The question and answer for each entry remain consistent across all groups.

\textbf{Position-Invariant Accuracy.} We select questions with four options from the multiple-choice question-answering dataset MMbench~\cite{Liu2023MMBenchIY}, referred to as the 4-option MMBench dataset. To calculate Position-Invariant Accuracy, for each question, we place the correct option in \textbf{\textit{Op.A}}, \textbf{\textit{Op.B}}, \textbf{\textit{Op.C}}, and \textbf{\textit{Op.D}} respectively to create four comparison groups.

\begin{figure}[t]
\centering
\includegraphics[width=0.45\textwidth]{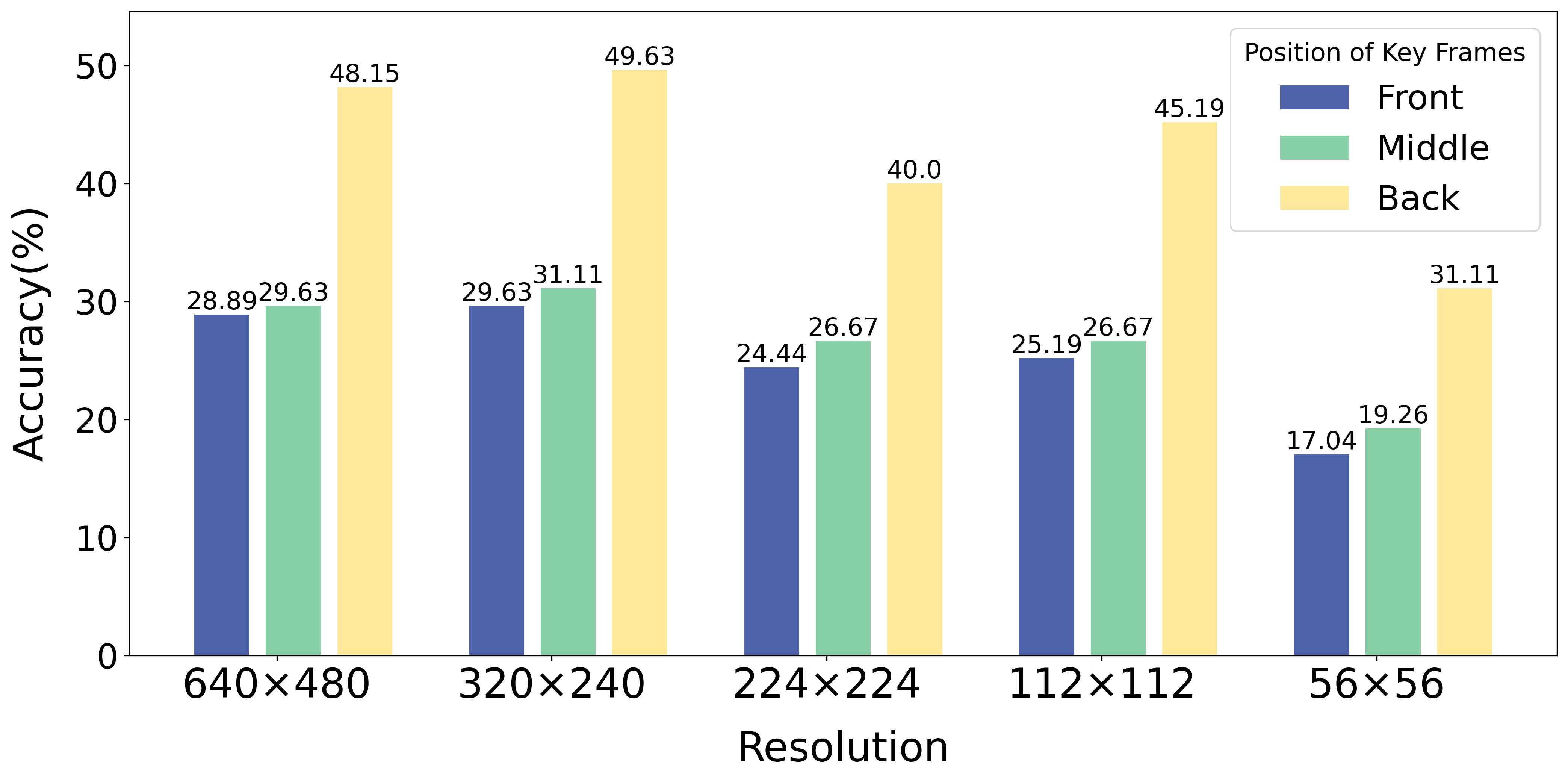} 
\caption{Statistics of correct responses for Video-LLaVA in video-caption matching task.}
\label{result1}
\end{figure}

\begin{figure}[t]
\centering
\includegraphics[width=0.42\textwidth]{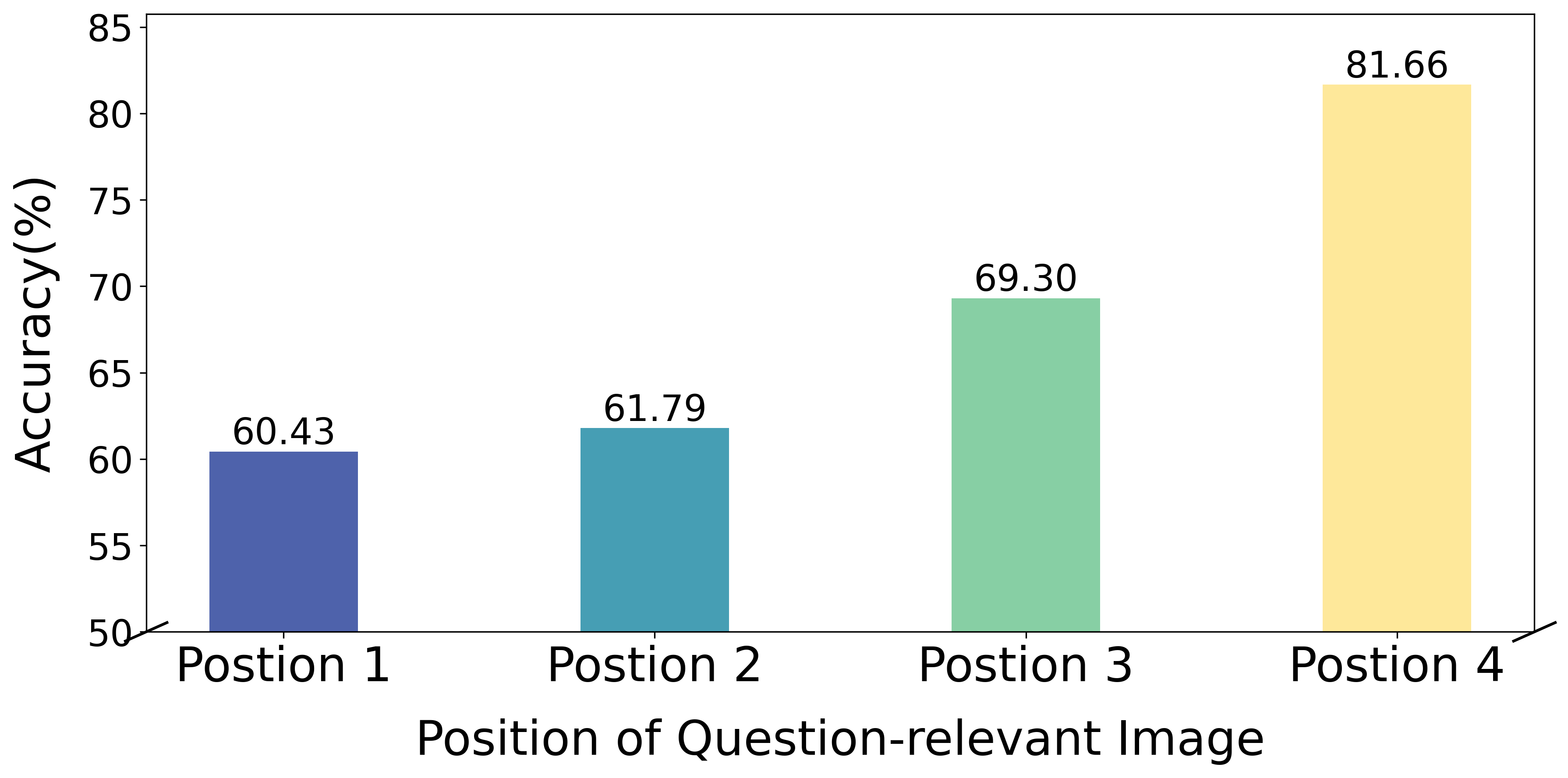} 
\caption{Statistics of correct responses for IDEFICS-v2-8B-Instruct in question-answering task with RAG.}
\label{result2}
\end{figure}

\begin{table*}[t]
    \centering
    \begin{tabular}{c|c|c|c|c}
    \toprule[1.5pt]
        Model & Parameters & Accuracy & Circular Evaluation & Position-Invariant \\
        & & (\%) & Accuracy (\%) & Accuracy (PIA) (\%) \\ \midrule
        InternLM-XComposer2~\cite{internlmxcomposer2} & 1.8B & 76.11 & 69.74 & 57.66 \\ 
        Qwen-VL-Chat~\cite{Qwen-VL} & 7B & 71.84 & 55.83 & 53.30 \\ 
        DeepSeek-VL~\cite{lu2024deepseekvl} & 7B & 80.13 & 73.89 & 64.24 \\ 
        IDEFICS-v2~\cite{Laurenon2024WhatMW} & 8B & 83.86 & 76.94 & 70.18 \\ 
        InternVL-Chat-v1.5~\cite{chen2023internvl} & 26B & 84.90 & 79.27 & 71.38 \\ 
        EMU2-Chat~\cite{Emu2} & 37B & 76.35 & 65.00 & 58.89 \\ \bottomrule[1.5pt]
    \end{tabular}
    \caption{Performance of different MLLMs on the 4-option MMBench dataset.}
    \label{PIA}
\end{table*}
\subsection{Results}
\textbf{Video-Caption Matching.} We set the frame sampling number to 15, ensuring the key action is sampled in all three comparative experiments. Figure~\ref{result1} shows the experimental results. Our independent repeated experiments conducted under 5 video resolutions ($640\times480, 320\times240, 224\times224, 112\times112,$ and $56\times56$) indicate that, regardless of the resolution, placing the \textit{Key Frames} at the back of the video yields the highest model accuracy, despite the same key action appearing in all three comparative experiments. Compared to positioning the \textit{Key Frames} at the beginning and middle, the average accuracy improves by 17.78\% and 16.15\% respectively across the 5 resolutions.

\textbf{Visual Question Answering with RAG.} Figure~\ref{result2} shows the experimental results. Although all four comparative groups contain the \textit{Question-relevant image}, positioning the \textit{Question-relevant image} in the last position of the RAG image set produces the highest model accuracy. Compared to the first three positions, the accuracy increases by 21.23\%, 19.87\%, and 12.36\%, respectively.

\textbf{Position-Invariant Accuracy.} Existing MLLM benchmarks overlook the model's sensitivity to order, where altering the correct answer's position can significantly impact model performance. Liu et al.~\cite{Liu2023MMBenchIY} introduce the Circular Evaluation Strategy to address this issue. In this approach, the correct answer is cyclically shifted across $M$ options and evaluated $M$ times, with the problem considered solved only if the model succeeds in all $M$ trials. However, this metric primarily assesses model robustness and fails to eliminate the inherent unfairness caused by order bias during evaluation. Consider a scenario where a model exhibits a strong tendency to select \textbf{\textit{Op.A}} with high frequency. In this case, choosing the correct option \textbf{\textit{Op.A}} might stem from this bias rather than true understanding, suggesting that a lower scoring weight should be applied to mitigate this tendency. Conversely, if the model rarely selects \textbf{\textit{Op.B}}, choosing the correct option \textbf{\textit{Op.B}} indicates high confidence and should be accorded a higher scoring weight.

To address these nuances, we propose a novel metric: Position-Invariant Accuracy (PIA). This metric is designed to account for and neutralize order biases in model responses, offering a more equitable evaluation of MLLM performance. The PIA is defined as follows:
\begin{equation}
    \text{PIA} = \frac{1}{M} \sum_{i=1}^{M} \frac{{C}_i}{{Pr}_i} \cdot \frac{{C}_i}{N},
\end{equation}
where $M$ represents the number of options, $C_i$ denotes the total number of times the model correctly answers the $i$-th option, $Pr_i$ represents the total number of times the model selects the $i$-th option, and $N$ represents the total number of questions in the dataset before constructing the comparison groups. $\frac{{C}_i}{{Pr}_i}$ indicates the extent to which the model's selection of option $i$ is due to its order bias. PIA assigns lower-scoring weights to the model's preferred positions and higher-scoring weights to its less preferred positions, thereby mitigating the unfairness caused by order bias in the evaluation results.

To demonstrate the effectiveness of PIA, as shown in Table~\ref{PIA}, we evaluate the accuracy, Circular Evaluation Accuracy, and Position-Invariant Accuracy (PIA) of 6 widely studied and used MLLMs on the 4-option MMBench dataset. The results show that PIA is smaller than both accuracy and  Circular Evaluation Accuracy. This is due to its scaling factor $\frac{{C}_i}{{Pr}_i}$, which penalizes the MLLM's tendency to blindly choose \textbf{\textit{Op.A}} or \textbf{\textit{Op.D}}.

\section{Future Work \& Conclusion}
The impact of order sensitivity on MLLM's performance is significant, with models often favoring the beginning and end of multimodal context. We believe this reflects human cognitive behavior - when the context is long, people tend to forget the middle content and focus on the beginning and end. The implications of this finding are twofold. Firstly, prompting techniques like Multi-Modal In-Context Learning (MMICL)~\cite{Khattak2022MaPLeMP, wang2024vilt,wu2024controlmllm} and Chain of Thought (CoT)~\cite{wei2022chain,zhang2023multimodal,mitra2024compositional} heavily rely on prompt order, making it necessary to re-examine the design of these tasks given the discovery of order sensitivity. Secondly, designing or training models also needs to pay more attention to order sensitivity. This is because current training data and task construction may inherently carry biases, with questions typically positioned at the beginning and end of context. This phenomenon calls for more diverse data and tasks.

In this paper, we demonstrate that MLLMs exhibit order sensitivity and find prompt orders that can improve performance. Popular MLLMs show a preference for the beginning and end of multimodal contexts, and placing important content in these positions can enhance performance. This is beneficial for improving accuracy in order-sensitive tasks such as video understanding and visual question answering. Finally, we propose a new MLLM metric: Position Invariant Accuracy (PIA), mitigating the unfair impact of order preferences. 

\appendix
\section{Appendix}
\subsection{A\quad Task Constructions on OpenFlamingo}

\begin{figure}[htbp]
\centering
\includegraphics[width=0.48\textwidth]{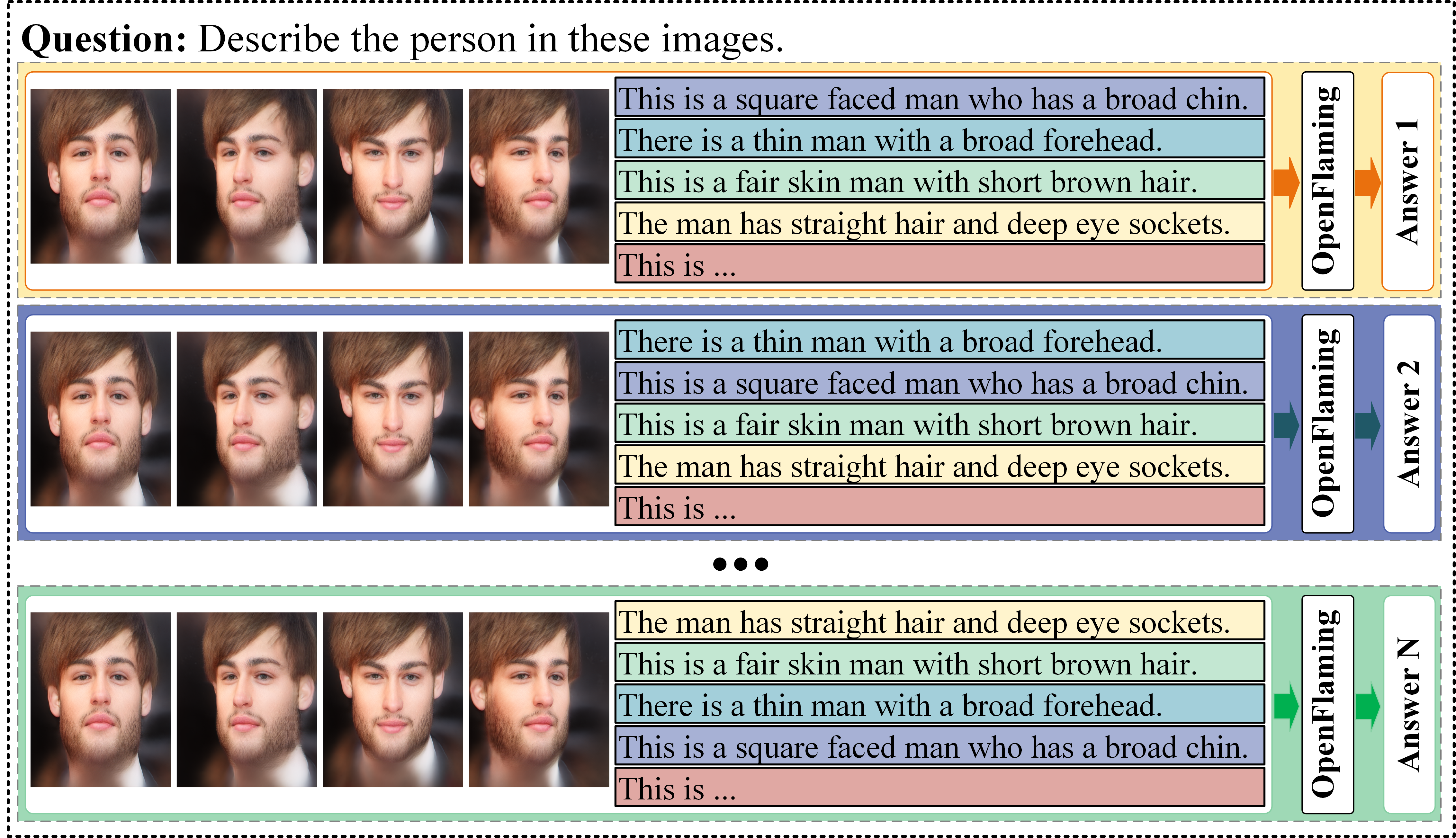} 
\caption{Description of text-only task on OpenFlamingo.}
\label{intro_text_only}
\end{figure}

\begin{figure}[htbp]
\centering
\includegraphics[width=0.48\textwidth]{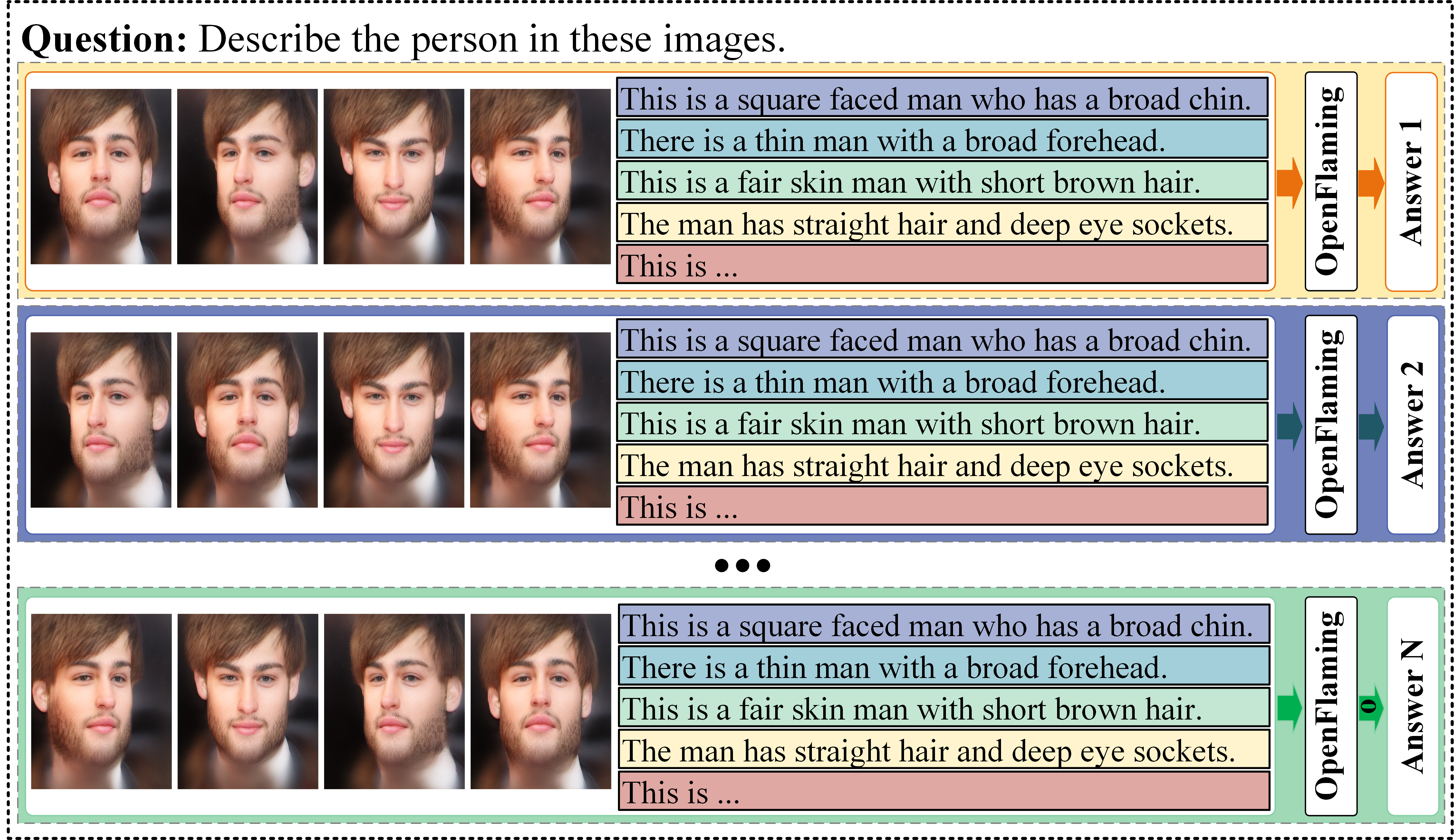} 
\caption{Description of image-only task on OpenFlamingo.}
\label{intro_image_only}
\end{figure}

Here we present text-only and image-only tasks constructed using the CelebAText-HQ dataset~\cite{Sun2021MulticaptionTS}. HFGI3D~\cite{Xie2022Highfidelity3G} is used to generate four distinct views of each facial image as visual prompts. For the text-only task, as shown in Figure~\ref{intro_text_only}, the order of facial image descriptions used as text prompts is altered in each experiment, while the order of facial images used as visual prompts remains constant. For the image-only task, as shown in Figure~\ref{intro_image_only}, conversely, the order of facial images used as visual prompts is changed in each experiment. In contrast, the order of facial image descriptions used as text prompts remains unchanged. The question is to describe the person in the images. OpenFlamingo~\cite{Awadalla2023OpenFlamingoAO} is asked to generate answers based on each prompt order.

\subsection{B\quad Task Constructions of Order Sensitivitty}
\begin{figure}[htbp]
\centering
\includegraphics[width=0.48\textwidth]{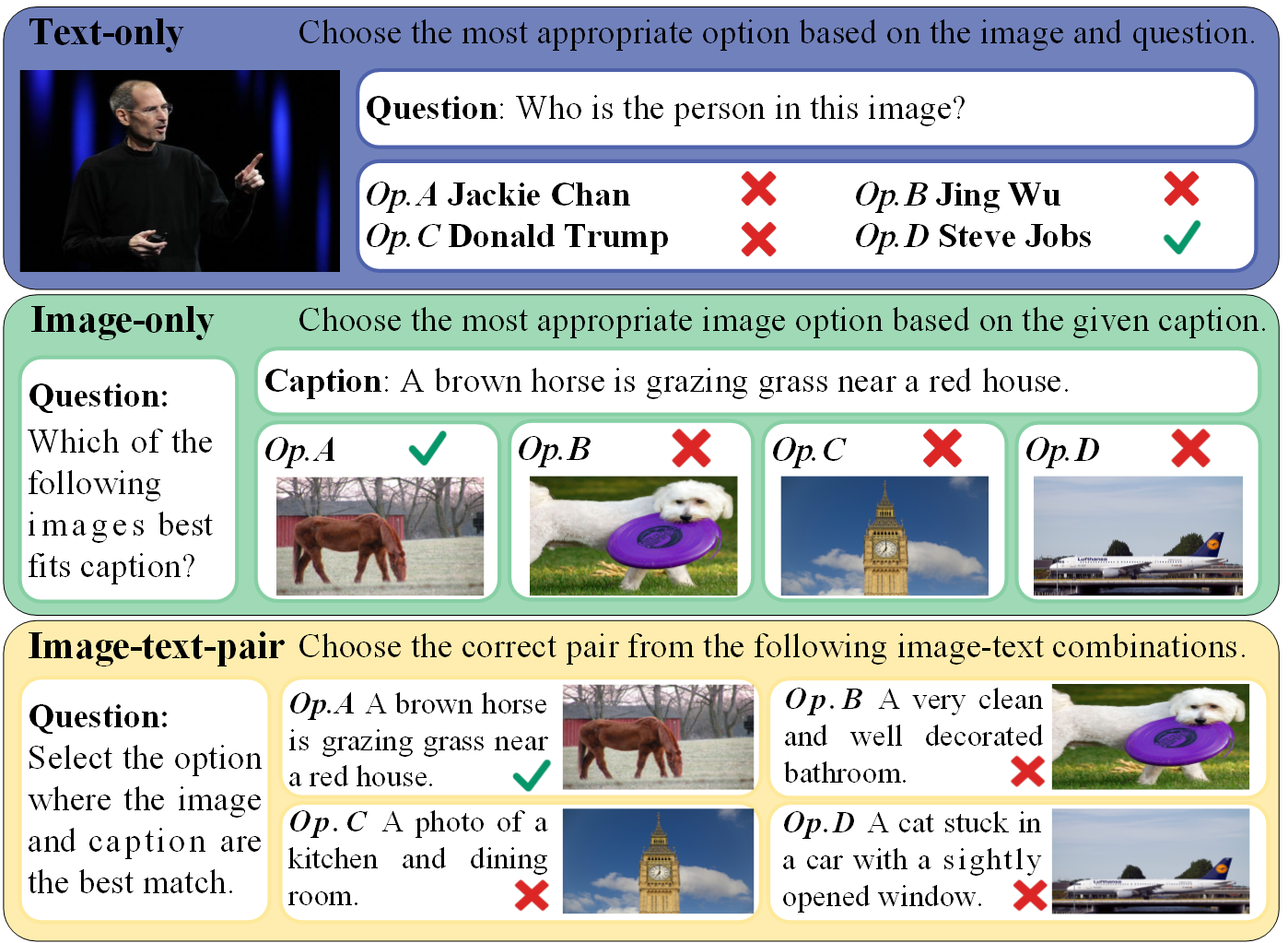} 
\caption{Tasks for order sensitivity.}
\label{table1}
\end{figure}

Here we present the details of task construction of order sensitivity. As shown in Figure~\ref{table1}. Text-only: Multiple-choice question-answering task, including a question related to an image and 4 text options. Image-only: Multiple-choice image-caption matching task, including a caption and 4 image options. Image-text-pair: Multiple-choice image-caption matching task, including 4 image-caption pair options.

\subsection{C\quad Setup \& Task Constructions of Experiment}
The experiment utilizes an AMD EPYC 7H12 64-core processor as the CPU and four NVIDIA RTX A6000 Ada GPUs. The system has 256 GB of memory. The operating system is Ubuntu 22.04.3. The experiments use CUDA version 12.4. The software environments for the experiments vary depending on the project; specific configurations can be found in our technical appendix. For experiments involving Video-LLaVA, Python version 3.10.4 and torch version 2.0.1 are used. For other experiments, Python version 3.9.12 and torch version 2.1.2 are employed.

\begin{figure*}[htbp]
\centering
\includegraphics[width=0.75\textwidth]{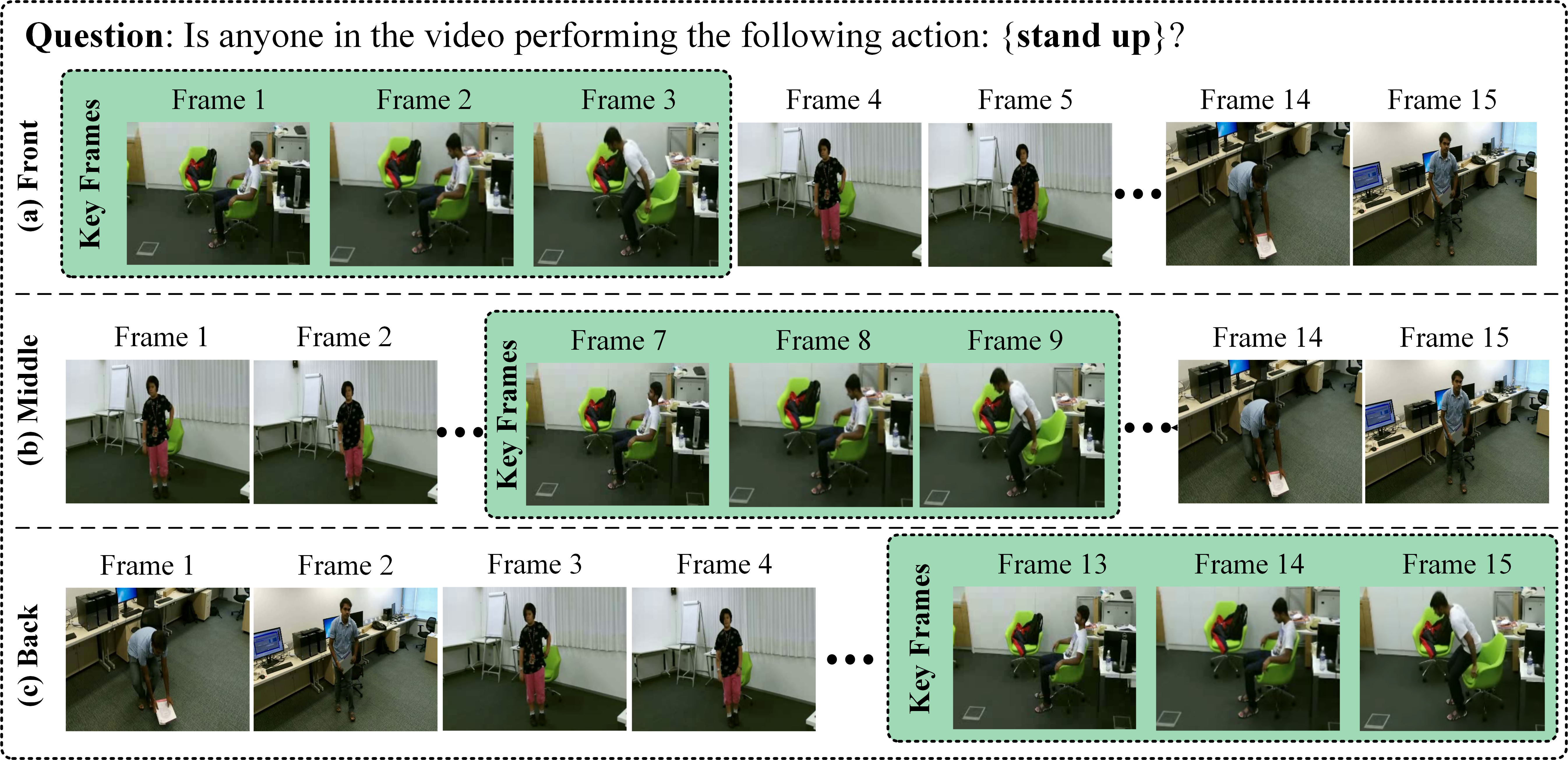} 
\caption{Demonstration of video-caption matching task.}
\label{exp_video}
\end{figure*}

We present the details of task construction of the video-caption matching task and visual question answering task with RAG. As shown in Figure~\ref{exp_video}, \textit{Key Frames} are placed at the front, middle, and back of the video, forming three comparative groups. The model is asked to answer questions separately. This allows us to verify whether the relative position of \textit{Key Frames} in the video affects performance. 

\begin{figure*}[htbp]
\centering
\includegraphics[width=0.75\textwidth]{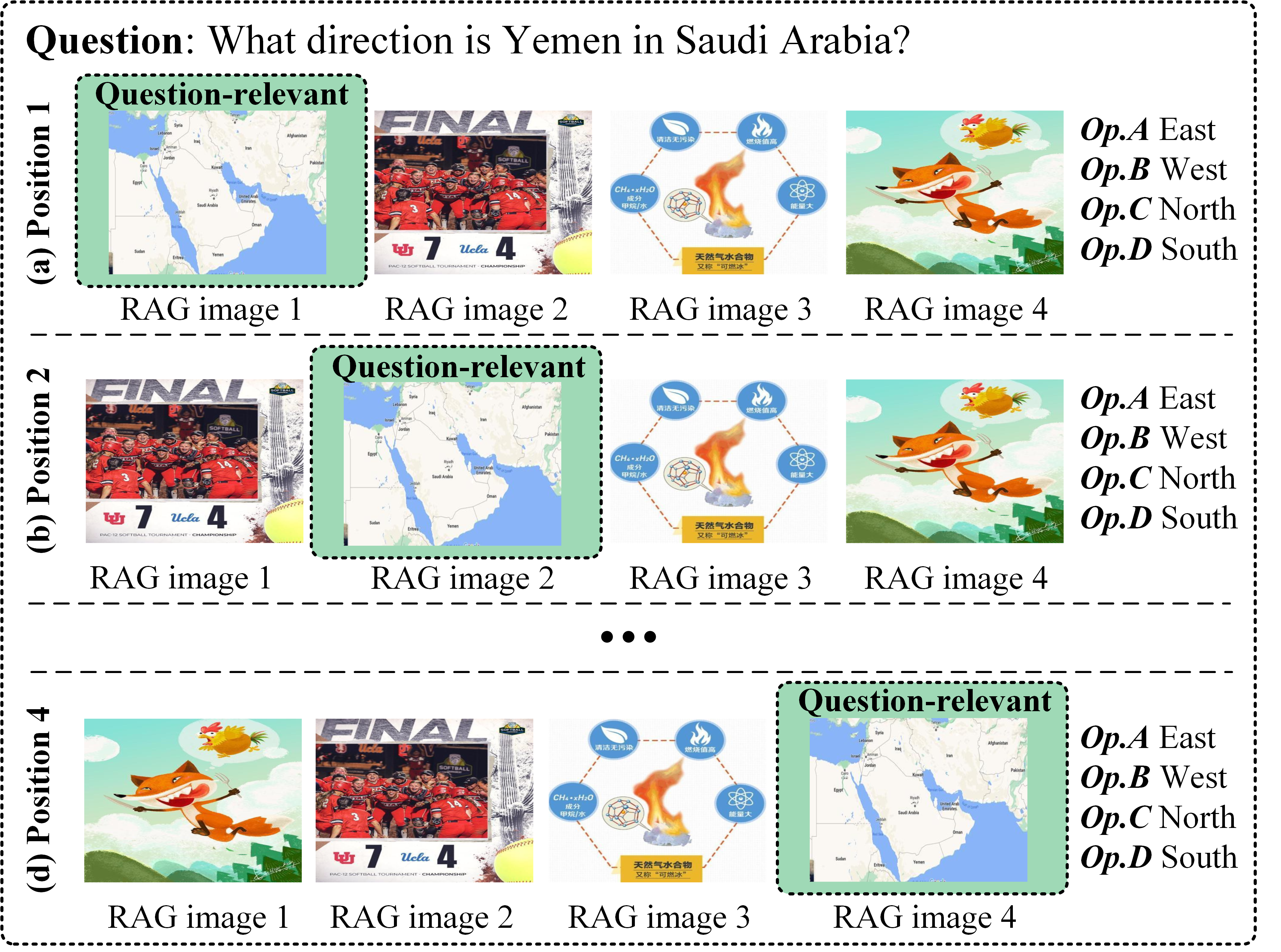} 
\caption{Demonstration of visual question answering task with RAG.}
\label{exp_rag}
\end{figure*}

As shown in Figure~\ref{exp_rag}, each question comes with a RAG image set containing 4 images, of which only one is the \textit{Question-relevant image}. The \textit{Question-relevant image} is placed in four different positions within the RAG image set, creating four comparative groups. The model is asked to answer questions separately. This allows us to verify whether the position of important content, i.e., the \textit{Question-relevant image}, in the prompts from RAG, affects performance.

\end{document}